\documentclass[10pt,twocolumn,letterpaper]{article}

\usepackage{cvpr}
\usepackage{booktabs}    
\usepackage{amssymb}     
\usepackage{colortbl}    
\usepackage{multicol}

\usepackage{url}            %
\usepackage{amsfonts}       %
\usepackage{nicefrac}       %
\usepackage{microtype}      %
\usepackage{xcolor}         %
\usepackage{marvosym}

\usepackage{pifont}
\newcommand{\cmark}{\ding{51}} 
\newcommand{\xmark}{\ding{55}} 
\usepackage{multirow} 
\usepackage{makecell}

\definecolor{cvprblue}{rgb}{0.21,0.49,0.74}
\usepackage[pagebackref,breaklinks,colorlinks,allcolors=cvprblue]{hyperref}

\title{Map-World: Masked Action planning and  Path-Integral World Model for Autonomous Driving}

\author{Bin Hu$^{1,* \diamond}$ \quad
Zijian Lu$^{2,*}$ \quad
Haicheng Liao$^{1}$ \quad
Chengran Yuan$^{2}$ \quad
Bin Rao$^{1}$ \quad
Yongkang Li$^{3}$ \quad \\
Guofa Li$^{4}$ \quad 
Zhiyong Cui$^{5}$ \quad
Cheng-zhong Xu$^{1}$ \quad
Zhenning Li$^{1~\textrm{\Letter}}$ 
\vspace{0.3em} \\
\quad\quad\quad \textsuperscript{1} University of Macau
\quad \textsuperscript{2} National University of Singapore
\quad \textsuperscript{3} Purdue University \\
\quad\quad\quad \textsuperscript{4} Chongqing University 
\quad \textsuperscript{5} Beihang University  \\}

\begin{document}
\maketitle
\let\thefootnote\relax\footnotetext{*: These authors contributed equally. $^\diamond$ Research Assistant of University of Macau; $^{~\textrm{\Letter}}$ Corresponding author: Zhenning Li(\url{zhenningli@um.edu.mo}).}
\begin{abstract}
Motion planning for autonomous driving must handle multiple plausible futures while remaining computationally efficient. Recent end-to-end systems and world-model-based planners predict rich multi-modal trajectories, but typically rely on handcrafted anchors or reinforcement learning to select a single “best” mode for training and control. This selection discards information about alternative futures and complicates optimization. We propose MAP-World, a prior-free multi-modal planning framework that couples masked action planning with a path-weighted world model. The Masked Action Planning (MAP) module treats future ego motion as masked sequence completion: past waypoints are encoded as visible tokens, future waypoints are represented as mask tokens, and a driving-intent path provides a coarse scaffold. A compact latent planning state is expanded into multiple trajectory queries with injected noise, yielding diverse, temporally consistent modes without anchor libraries or teacher policies. A lightweight world model then rolls out future BEV semantics conditioned on each candidate trajectory. During training, semantic losses are computed as an expectation over modes, using trajectory probabilities as discrete path weights, so the planner learns from the full distribution of plausible futures instead of a single selected path. On NAVSIM, our method matches anchor-based approaches and achieves state-of-the-art performance among world-model-based methods,while avoiding reinforcement learning and maintaining real-time inference latency.

\end{abstract}    
\section{Introduction}
\label{sec:intro}

End-to-end autonomous driving maps raw multi-sensor inputs to future ego trajectories or low-level controls within a single network~\cite{chitta2022transfuser, li2024law,hu2023_uniad}, building on recent advances in 3D perception, motion forecasting, and online mapping~\cite{polar, huang2022bevdet, wang2021detr3d,li2022bevformer,shi2023motiontransformer,liao2023maptr}. This paradigm reduces hand-crafted interfaces and has reached strong performance on several open- and closed-loop benchmarks. Yet a central difficulty remains: in realistic traffic, there are many plausible futures for both the ego vehicle and surrounding agents, and the planner must represent this multi-modality without sacrificing temporal consistency or real-time efficiency.

\begin{figure}[!]
    \centering
    \includegraphics[width=\linewidth]{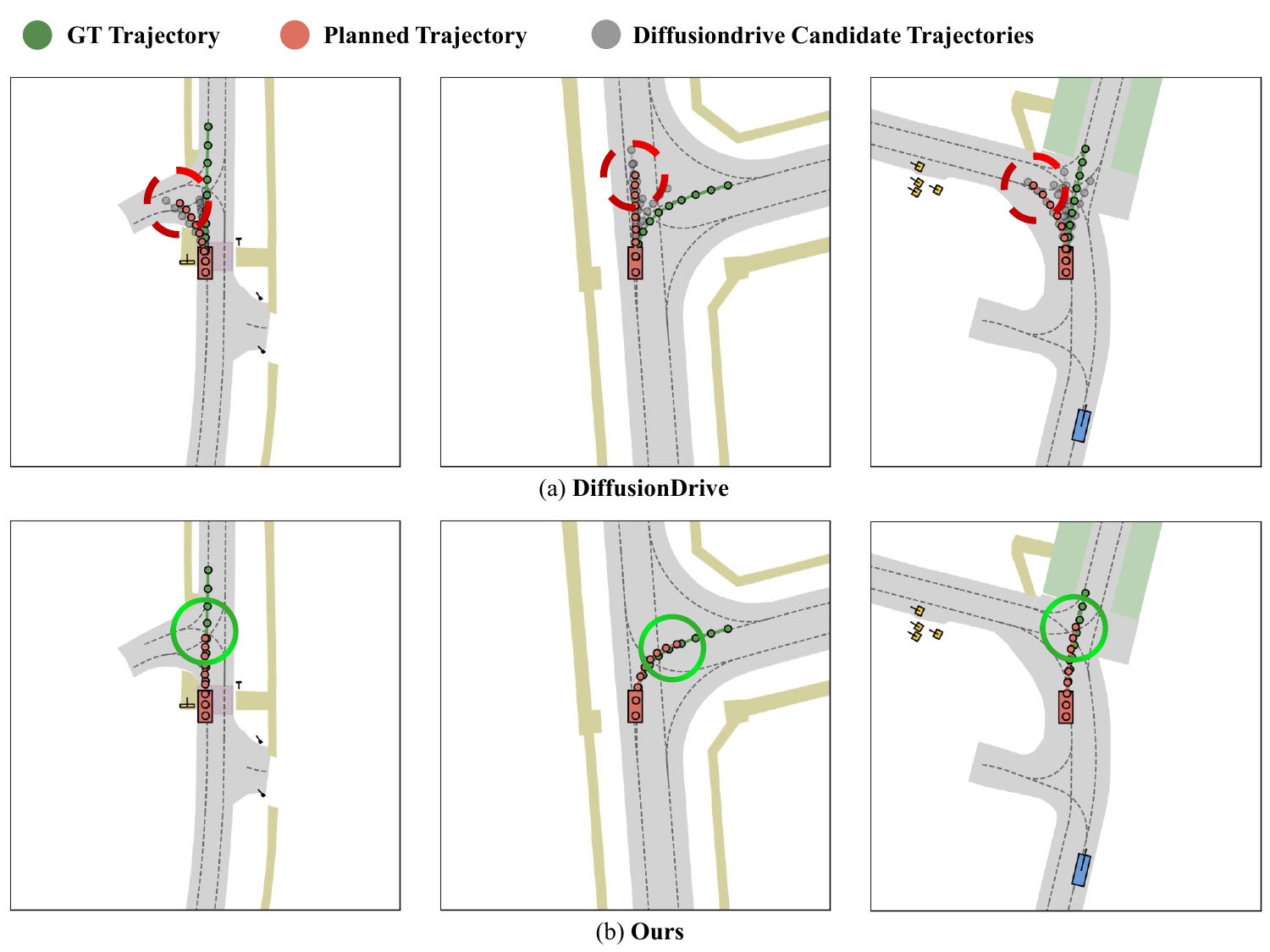}
    \caption{Anchor-based selection versus MAP-World. (a) DiffusionDrive generates multi-modal trajectories tied to an anchor set and then selects one as the final plan. (b) MAP-World predicts trajectories directly via masked action planning, without anchors, allowing a broader family of motion modes that better aligns with the ground truth.}
    \label{fig:fig1}
\end{figure}

Most current systems handle multi-modality through a fixed set of trajectory anchors. The network learns to refine or select from this library~\cite{diffusiondrive, li2025wote, zheng2025world4driveendtoendautonomousdriving, li2024hydra}, which stabilizes training and avoids trivial mode collapse. However, anchors discretize the solution space. Important behaviours that depend on subtle interactions or road geometry may fall between anchors, leading to plans that are feasible but misaligned with natural driving and sometimes inconsistent with the observed history. The matching and selection stages also add latency and implementation complexity, and even reduced-anchor designs such as DiffusionDrive~\cite{diffusiondrive} remain constrained by the expressiveness of the anchor set (Fig.~\ref{fig:fig1}).

World-model-based planners aim to improve foresight by predicting how the scene will evolve. Some methods first roll out future BEV states and then regress trajectories from these predictions~\cite{li2024law,yang2025resim, zheng2023occworld}; others generate candidates and rely on a learned world model to score or filter them, often with a reinforcement-learning selector~\cite{li2025wote, zheng2025world4driveendtoendautonomousdriving}. Both families enrich the planner with counterfactual rollouts, but they largely follow a “generate then pick one’’ pattern. Only the selected trajectory influences control and supervision, so most of the multi-modal structure is discarded, and selector training introduces additional complexity.

This work asks whether the planner and world model can be coupled in a different way: can we generate multi-modal futures without anchors, keep them consistent with history and intent, and train the world model on the full distribution of plausible paths rather than on a single choice?

We answer this with \textbf{MAP-World}, a prior-free multi-modal planning framework that combines masked action planning with path-weighted world-model training. Future ego motion is treated as a masked sequence to be completed: past waypoints are visible tokens, future waypoints are mask tokens, and a driving-intent trajectory provides a coarse scaffold. From this sequence and the BEV scene representation, MAP-World constructs a compact latent planning state. Injecting noise into this state yields multiple trajectory queries that share the same history and intent but diverge in their futures, so multi-modality arises from a learned latent space rather than from an anchor library. A lightweight world model predicts future BEV semantics conditioned on each trajectory, and a trajectory probability head defines a distribution over modes. Training minimizes a semantic loss that is an expectation over trajectories, weighted by their probabilities, so the planner learns from an ensemble of plausible paths while remaining fully differentiable and free of reinforcement learning.

In summary, our contributions are:
\begin{itemize}
    \item We propose a prior-free multi-modal trajectory generator that formulates planning as masked sequence completion (visible past, masked future, intent scaffold), producing diverse and history-consistent trajectories without anchors or teacher policies.

    \item We develop a path-weighted world-model objective that supervises an expectation over candidate trajectories using their predicted probabilities as importance weights, enabling end-to-end differentiable training without reinforcement learning or post-hoc selection.

    \item Extensive experiments show that MAP-World achieves state-of-the-art closed-loop results on NAVSIM and competitive open-loop accuracy on nuScenes, while keeping inference latency compatible with real-time deployment.
\end{itemize}
\section{Related Works}
\label{sec:Related}
\subsection{End-to-End Autonomous Driving}
End-to-end driving maps raw sensor inputs to waypoints or controls within a single model \cite{centaur,hu2023_uniad,li2024law,yuan2024drama,sun2024sparsedriveendtoendautonomousdriving}. UniAD \cite{hu2023_uniad} unifies perception to strengthen planning. VAD \cite{jiang2023vad} vectorizes agents and HD maps into a compact scene representation, improving planning efficiency. VAD-V2 \cite{chen2024vadv2} introduces vectorized scene representations and large-vocabulary probabilistic planning for multi-modal trajectories, but offers limited guarantees of diversity and quality. HydraMDP \cite{li2024hydra} combines imitation and reinforcement learning with a teacher planner, increasing dependence on prior design and training complexity. DiffusionDrive \cite{diffusiondrive} uses diffusion with anchor trajectories and truncated denoising for real-time inference. MomAD \cite{song2025momad} injects historical information as trajectory and perceptual momentum into current decisions, mitigating jitter and myopia from weak temporal modeling. Despite these advances  \cite{chen2024vadv2,diffusiondrive,li2025wote,song2025momad,zheng2025world4driveendtoendautonomousdriving,liu2025gaussianfusion,xing2025goalflow,chai2025anchdrive,yao2025drivesuprim,zhang2025perceptionplan}, many methods still rely on predefined priors and auxiliary selection modules, which constrain diversity, tie quality to prior design, and increase system overhead.

\subsection{World Model in Autonomous Driving}
LAW \cite{li2024law} divides driving world models into image-based approaches \cite{wang2023drivingwm,yang2025resim,hu2023gaia1,hu2022urban} that generate future images and then plan, and occupancy-based approaches \cite{zheng2023occworld,min2024driveworld} that forecast future occupancy and then plan. The former is computation-heavy, and the latter requires high-quality occupancy labels. Both adopt a predict-then-plan pipeline with high latency. LAW instead plans first, then predicts future states conditioned on the plan and supervises them, improving trajectory quality. Building on this idea, WoTE   \cite{li2025wote} pairs BEV features with trajectory anchors and uses RL to select plausible futures, and World4Drive  \cite{zheng2025world4driveendtoendautonomousdriving} extends LAW to multi-modal trajectories via pretrained intent cues,yet both still rely on predefined anchors for multi-modality.

\subsection{Masked AutoEncoder in Autonomous Driving}
Occupancy-MAE \cite{min2023occmae} adapts MAE \cite{MaskedAutoencoders2021} to voxelized LiDAR by masking and reconstructing occupancy, reducing label requirements and improving downstream performance. Traj-MAE \cite{chen2023trajmae} reconstructs masked histories and HD-map elements to learn interaction-aware features. M-BEV \cite{chen2023mbev} applies MAE to camera-to-BEV perception via view-occlusion reconstruction. UniM$^{2}$AE \cite{zou2023unimae} unifies images and LiDAR in a shared 3D volume. And NOMAE \cite{abdelsamad2025nomae} restricts masking to neighborhoods of occupied voxels for more efficient self-supervised learning. However, these MAE variants primarily reconstruct observed data rather than forecast future states; to our knowledge, this is the first MAE-based extension to a latent world model for autonomous driving.

\section{Method}
\label{sec:Method}
\begin{figure*}[t!]
\centering
\includegraphics[width=1\textwidth]{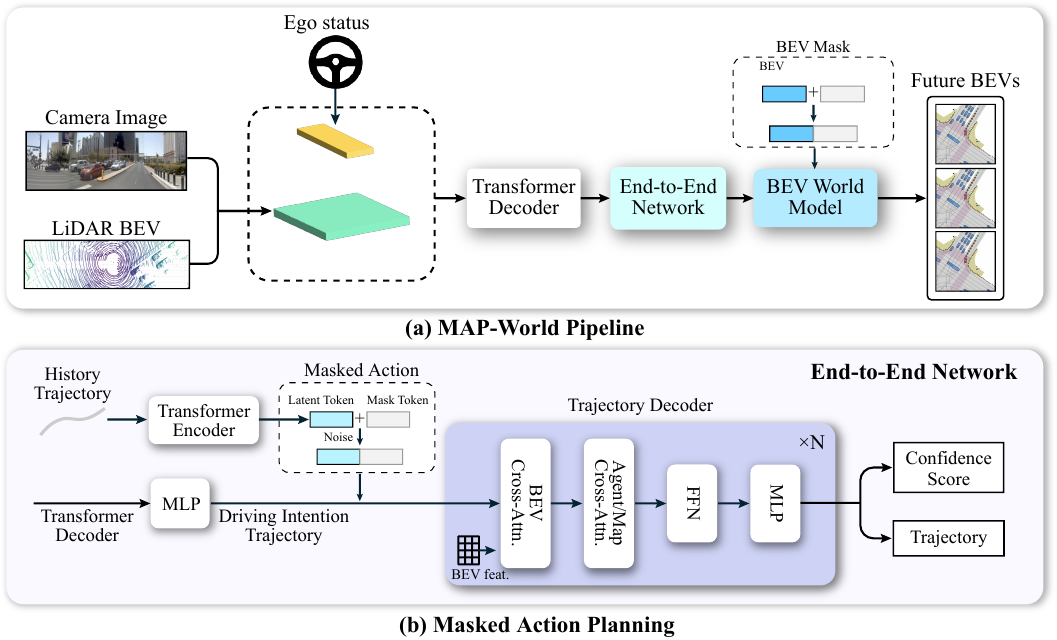}
\caption{
Overview of MAP-World. (a) Multi-view images and LiDAR are encoded to obtain the current BEV features. The encoded ego state is fused with the BEV features to form the current state representation. (b) Masked Action Planning generates multi-modal trajectories by applying a Transformer decoder to the current state representation. (c) The BEV world model conditions on the multi-modal trajectories and current BEV features to synthesize future BEV features, which are trained via losses against the BEV semantic map and evaluated under a path-integral formulation.
}
\label{fig:main}
\end{figure*}

\subsection{Preliminary}
\noindent\textbf{Task formulation.} End-to-end autonomous driving maps raw sensor inputs to a scene representation and, conditioned on this representation, predicts the ego vehicle’s future trajectory. The trajectory is represented as a sequence of waypoints\(\mathbf{T}_t = \{\mathbf{\tau}_t^1, \mathbf{\tau}_t^2, \ldots, \mathbf{\tau}_t^L\}\), where each waypoint \(\mathbf{\tau}_t^i = (x_t^i, y_t^i)\) denotes the predicted BEV position of the ego vehicle at time \(t + i\). The horizon \(L\) specifies the number of future positions to be predicted. 

\noindent\textbf{World model.} A world model predicts the world features or state at time \(t+i\) from those at time \(t\). In autonomous driving, such models either condition on current images or BEV features and forecast future visual or BEV representations to guide trajectory planning, or condition on current BEV features together with a predicted trajectory and jointly forecast future BEV features to further improve planning performance.

\noindent\textbf{Masked autoencoder.} It \cite{MaskedAutoencoders2021} employs a high masking ratio and reconstructs the masked patches with an asymmetric architecture to learn globally coherent representations. This reconstruction-based pretraining substantially reduces computational cost and improves scalability and transferability.

\noindent\textbf{Feynman path-integral.} Feynman’s path integral formulates quantum evolution as a sum over all spacetime paths satisfying endpoint constraints, with each path contributing to the transition amplitude via a phase weight. Thus “all histories interfere.” A compact expression in terms of a path measure and phase weight is:
\begin{equation}
\begin{aligned}
Z \;&=\; \int \mathcal D\tau \;\exp\!\big(\mathcal A[\tau]\big) \\[4pt]
\mathcal A[\tau] \;&=\; \int \ell\!\big(\tau(s),\tau'(s),t\big)\,dt
\end{aligned}
\label{eq:feynman}
\end{equation}
\(D\tau\) denotes the path-integral measure, i.e., integration over all admissible paths. \(A[\tau]\) denotes the action functional, integrating the Lagrangian over time yields the action. \(\tau(s)\) represents a path, and \(\tau'(s)\) its derivative with respect to \(t\), which can be interpreted as a velocity or rate of change.
\subsection{Masked Action Planning}
In this section, we plan future trajectories using the historical trajectory and the current BEV state. We first derive an intent trajectory from the current BEV features. Following the MAE paradigm, the historical trajectory is encoded as unmasked latent tokens, while the future trajectory is represented by masked tokens. A conditional generative decoder then reconstructs multi-modal future trajectories from these tokens.

\noindent\textbf{Driving intention trajectory.} We define a driving-intention trajectory as an initial hypothesis of the ego vehicle’s future intent inferred from the current state, serving as a coarse proxy for the downstream route and path geometry. Concretely, we adopt TransFuser\cite{chitta2022transfuser} as the perception–fusion backbone to integrate camera images and LiDAR into a BEV feature map \(F_{bev}\). Following the prior work \cite{diffusiondrive, li2025wote, li2024law}, we concatenate these BEV tokens with an ego status embedding \(Emb_{ego}\) produced by a dedicated linear layer to obtain current BEV state \(F^{cur}_{state}\), and apply learnable queries \(Q_{bev}\) with cross-attention over the BEV representation to disentangle ego intention \(F_{ego}\) and agent intention \(F_{Agent}\) features. Finally, an MLP projects the \(F_{ego}\) to a driving-intent trajectory parameterized in the current ego- centric coordinate system. Given the 
multi-view image \(I\) and LiDAR feature \(Li\),we can obtain:
\begin{equation}
\begin{aligned}[c]
F_{bev} &= TransFuser(I,Li) \\
F^{cur}_{state} &= concat(F_{bev}, Emb_{ego}) \\
F_{ego}, F_{Agent} &= CrossAttention(Q_{bev},F^{cur}_{state}) \\
T_{intent} &= MLP(F_{ego})
\end{aligned}
\label{eq:coarse}
\end{equation}
We then employ a lightweight BEV semantic decoder to map BEV features into a BEV semantic map used for supervision. In parallel, a separate MLP projects agent representations to agent states \(\mathcal{S}_{Agent}\) and computes a loss \(\mathcal{L}_{Agent}\) against the agent ground-truth.
\begin{equation}
\begin{aligned}[c]
\mathcal{B}_{semantic} &= BEVdecoder(F_{bev}) \\
\mathcal{S}_{Agent} &= MLP(F_{Agent})
\end{aligned}
\label{eq:coarse}
\end{equation}

\noindent\textbf{Trajectory decoder.}Given a history of length \(T_h\) waypoints  \(\{\tau_t\}_{t=1}^{T_h},\tau_t \in \mathbb{R}^2\), the module constructs a MAE- style decoder that reconstructs multi-modal future trajectories conditioned on BEV context and auxiliary cues.

The history is augmented with first-order motion cues by concatenating displacements \(\Delta\mathbf{\tau}_t=\mathbf{\tau}_t-\mathbf{\tau}_{t-1}\), yielding \([\mathbf{\tau}_t,\Delta\mathbf{\tau}_t]\in\mathbb{R}^4\). An MLP projects each step to a d-dimensional token, and a Transformer encoder produces temporally contextualized history embeddings:
\begin{equation}
\begin{aligned}[c]
F_{hist} &= \mathrm{MLP}\big([\mathbf{\tau}_1,\Delta\mathbf{\tau}_1],\dots,[\mathbf{\tau}_{T_h},\Delta\mathbf{\tau}_{T_h}]\big)\in\mathbb{R}^{T_h\times d} \\
\mathbf{H}&=\mathrm{Encoder(F_{hist})}
\end{aligned}
\label{eq:hist}
\end{equation}
Future steps \(T_f\) are represented by learnable mask tokens. The decoder input is formed by concatenating the encoded history tokens with these future mask tokens:
\begin{equation}
\begin{aligned}[c]
\mathbf{Q_{traj}} = [\mathbf{H}, \mathbf{Mask}] \in \mathbb{R}^{(T_h + T_f)\times d}
\end{aligned}
\label{eq:hist}
\end{equation}
To realize multi-modal futures, the sequence \(\mathbf{Q_{traj}}\) is compressed across time to \(\tilde{q}\in \mathbb{R}^d \) by a fusion MLP and replicated to \(K\) mode query \(\{\tilde{q}^{(k)}\}_{k=1}^{K}\). Furthermore, per-mode latent noise is mapped and added to induce diverse hypotheses:
\begin{equation}
\begin{aligned}[c]
\mathbf{q}^{(k)}=\tilde{\mathbf{q}}+\psi(\mathbf{z}^{(k)}),
\mathbf{z}^{(k)}\sim\mathcal{N}(\mathbf{0},\mathbf{I})
\end{aligned}
\label{eq:hist}
\end{equation}
The full  coordinate sequence \(\mathbf{P}\in\mathbb{R}^{(T_h+T_f)\times 2}\) is formed by concatenating history waypoints \(\{\tau_t\}_{t=1}^{T_h}\) with a driving intent trajectory \(T_{intent}\). For each mode k, the decoder receives the context feature \(F_{context} = \{\mathbf{q}^{(k)},\mathbf{P},  F_{bev}, F_{Agent}\}\). Within the trajectory decoder, these conditions are fused via cross-attention.\(F_{bev}\) supplies spatial context, \(\mathbf{P}\) imposes geometric constraints, and auxiliary signals provide semantic and dynamical priors. The decoder predicts each mode’s trajectory as a residual relative to the driving-intent trajectory, thereby refining mode-specific representations and producing the final candidate set \(T_{Tf}\) together with the corresponding path probabilities or confidence score \(T_{cls}\):
\begin{equation}
\begin{aligned}[c]
F_{scene}&=TrajDecoder(F_{context}) \\ 
T_{res}, T_{cls} &= RefineModule(F_{scene}) \\
T_{Tf} &= T_{res} + \mathbf{P}
\end{aligned}
\label{eq:hist}
\end{equation}
Follow the prior work\cite{diffusiondrive},the trajectory decoder first interacts with BEV features via deformable spatial cross-attention. The resulting trajectory features then perform cross-attention with the agent feature \(F_{Agent}\), followed by a feed-forward network(FFN). A final MLP refinement head estimates the confidence of each trajectory and its offset with respect to the reference path \(\mathbf{P}\). The predicted confidence is interpreted as the path weight in the Feynman path–integral formulation and is also used for trajectory selection. We take the trajectory with the highest confidence as the final prediction.

Masked Action Planning encodes the observed history and treats the future as masked targets. A conditional Transformer decoder, conditioned on BEV context, agent tokens, and a coarse intent path, reconstructs the future trajectory. Multi-modality arises from mode queries with latent perturbations, jointly optimizing geometric fidelity and inter-mode separation. The history supplies dynamics priors and boundary conditions, while the masked-sequence formulation enforces temporal consistency and reduces drift. Predicting residuals with respect to the intent trajectory further improves synthesis quality. We use $\ell_1$ loss for trajectories, focal loss for trajectory classification, cross-entropy loss for BEV semantics, and a combination of cross-entropy and $\ell_1$ for agent boxes and labels, aggregated as:

\begin{equation}
\begin{aligned}[c]
\mathcal{L}_{e2e} &= \lambda_{traj}\mathcal{L}_{traj}(T_{Tf}, T_{gt})\\ &+\lambda_{agent}\mathcal{L}_{Agent} + \lambda_{semantic}\mathcal{L}_{semantic} \\
&+\lambda_{cls}\mathcal{L}_{cls}
\end{aligned}
\label{eq:losse2e}
\end{equation}
\subsection{Feynman Path Integral World Model}
\noindent\textbf{Latent world model.}
Following prior work \cite{li2024law, li2025wote},the latent world model takes as input the current BEV features and a candidate future trajectory, and predicts future BEV features. We also attach an MAE-style masked reconstruction head: learnable mask tokens with positional embeddings are concatenated with the current BEV tokens and decoded into future BEV latents. This decouples current and future representations and biases the model toward generating future semantics rather than copying the present. Compared with directly regressing future BEV features from current BEV and trajectory embeddings, this design yields a clearer optimization objective and reduces interference between geometric reconstruction and temporal extrapolation.

\noindent\textbf{Path-integral view.} We consider discrete time indices $t=1,\dots,T$ with a history horizon $T_h$ and a prediction horizon $T_f$ so that $T=T_h+T_f$.
We condition on the observed history, a driving intention path, and the current BEV state, the conditioning set:
\[
\mathcal{C}
\;\triangleq\;
\Big\{
\boldsymbol{\tau}_{1:T_h},\;
\mathrm{T_{intent}},\;
F^{\mathrm{cur}}_{\mathrm{state}},\;
F_{\mathrm{Agent}}
\Big\},
\]
Integrate only over the \emph{future} degrees of freedom and future trajectory-field representation follows as:
\[
\Phi
\;\triangleq\;
\Big\{
\boldsymbol{\tau}_{T_h+1:T_h+T_f},\;
\mathcal{B}_{\mathrm{semantic}}^{(T_h+1:T_h+T_f)}
\Big\},
\]
The conditional law over future configurations given $\mathcal{C}$ is written as
\begin{equation}
\begin{aligned}[c]
\mathcal{Z}[\mathcal{C}]
\;&=\;
\int \mathcal{D}\Phi\;
\exp\!\big(-\mathcal{A}[\Phi;\mathcal{C}]\big) \\
\mathcal{D}\Phi
\;&\triangleq\;
\prod_{t=T_h+1}^{T_h+T_f}
\mathrm{d}\boldsymbol{\tau}_{t}\;\mathrm{d}\mathcal{B}_{\mathrm{semantic}}^{(t)}.\\
&\iff T_{cls}
\end{aligned}
\label{eq:conditionlaw}
\end{equation}
with the discrete-time action decomposed as
\begin{equation}
\label{eq:S-discrete-mine-en}
\begin{aligned}[c]
\mathcal{A}[\Phi;\mathcal{C}]
&=
\sum_{t=T_h+1}^{T_h+T_f}
\Big(
\lambda_{\mathrm{bev}}\,
\ell_{\mathrm{bev}}\!\big(\mathcal{B}_{\mathrm{semantic}}^{(t)}\;\big|\;F^{\mathrm{cur}}_{\mathrm{state}}\big)\Big) \\
&+\mathcal{B}[\mathcal{C}], \\
&\iff \mathcal{L}_{wm}(WorldModel(T_{Tf}, F_{bev}))
\end{aligned}
\end{equation}
where $\mathcal{B}[\mathcal{C}]$ is a boundary term that enforces the endpoint constraints:
\begin{equation}
\label{eq:boundary}
\begin{aligned}[c]
\mathcal{B}[\mathcal{C}] = &-\log \delta\!\big(\boldsymbol{\tau}_{1:T_h}-\boldsymbol{\tau}^{\,\mathrm{obs}}_{1:T_h}\big) \;\\&
-\log \delta\!\big(F^{\mathrm{cur}}_{\mathrm{state}}-\mathrm{concat}(F_{bev},Emb_{ego})\big).
\end{aligned}
\end{equation}
The boundary term encodes hard endpoint constraints via Dirac deltas,
\(\delta(\boldsymbol{\tau}_{1:T_h}-\boldsymbol{\tau}^{\mathrm{obs}}_{1:T_h})\) clamps the historical trajectory to observations, and \(\delta(F^{\mathrm{cur}}_{\mathrm{state}}-\mathrm{concat}(F_{bev},Emb_{ego}))\) anchors the current BEV state. Expressed as \(-\log\delta(\cdot)\), the penalty is zero when satisfied and \(+\infty\) otherwise, implying these variables are fixed conditioning rather than optimized degrees of freedom. Note that this term merely indicates that these components do not belong to the future degrees of freedom to be integrated, and it is unrelated to our actual training loss.

In Masked Action Planning, the historical segment is observed, and the\(T_f\)future steps are represented by learnable mask tokens. In a path-integral view, this specifies boundary conditions (fixed past, integrable future),the mask tokens act as placeholders for future degrees of freedom, whose posteriors are reconstructed by the decoder under the conditioning. The corresponding path weight is the selection probability of each trajectory among the generated multi-modal candidates.Therefore, finally, we can express the Feynman path integral in the world model as follows:
\begin{equation}
\label{eq:overall-fpi}
\begin{aligned}[c]
\mathcal{Z}[\mathcal{C}]
\;&=\;
\int \mathcal{D}\Phi\;
\exp\!\big(-\mathcal{A}[\Phi;\mathcal{C}]\big) \\
&\iff\sum_{k=1}^KT_{cls}^{k} \mathcal{L}_{wm}(\hat{\mathcal{B}}_{semantic}^{(Tf)}\
,\mathcal{B}_{semantic}^{(Tf)})
\end{aligned}
\end{equation}
where \(\hat{\mathcal{B}}_{semantic}^{(Tf)}=\textit{WorldModel} (T_{Tf}, F_{bev})\),
\(\mathcal{B}_{semantic}^{(T_f)}\) means ground-truth semantic map at future time \textit{Tf}. CE loss is also used here for the semantic map. Therefore, the total loss:

\begin{equation}
\label{eq:totalloss}
\begin{aligned}[c]
\mathcal{L}_{total} = \mathcal{L}_{e2e} + \lambda_{wm}
\sum_{k=1}^KT_{cls}^{k} \mathcal{L}_{wm}(\hat{\mathcal{B}}_{semantic}^{(Tf)}\
,\mathcal{B}_{semantic}^{(Tf)})
\end{aligned}
\end{equation}

\section{Experiments}
\label{sec:Experiments}
\begin{figure*}[t!]
\centering
\includegraphics[width=1\textwidth]{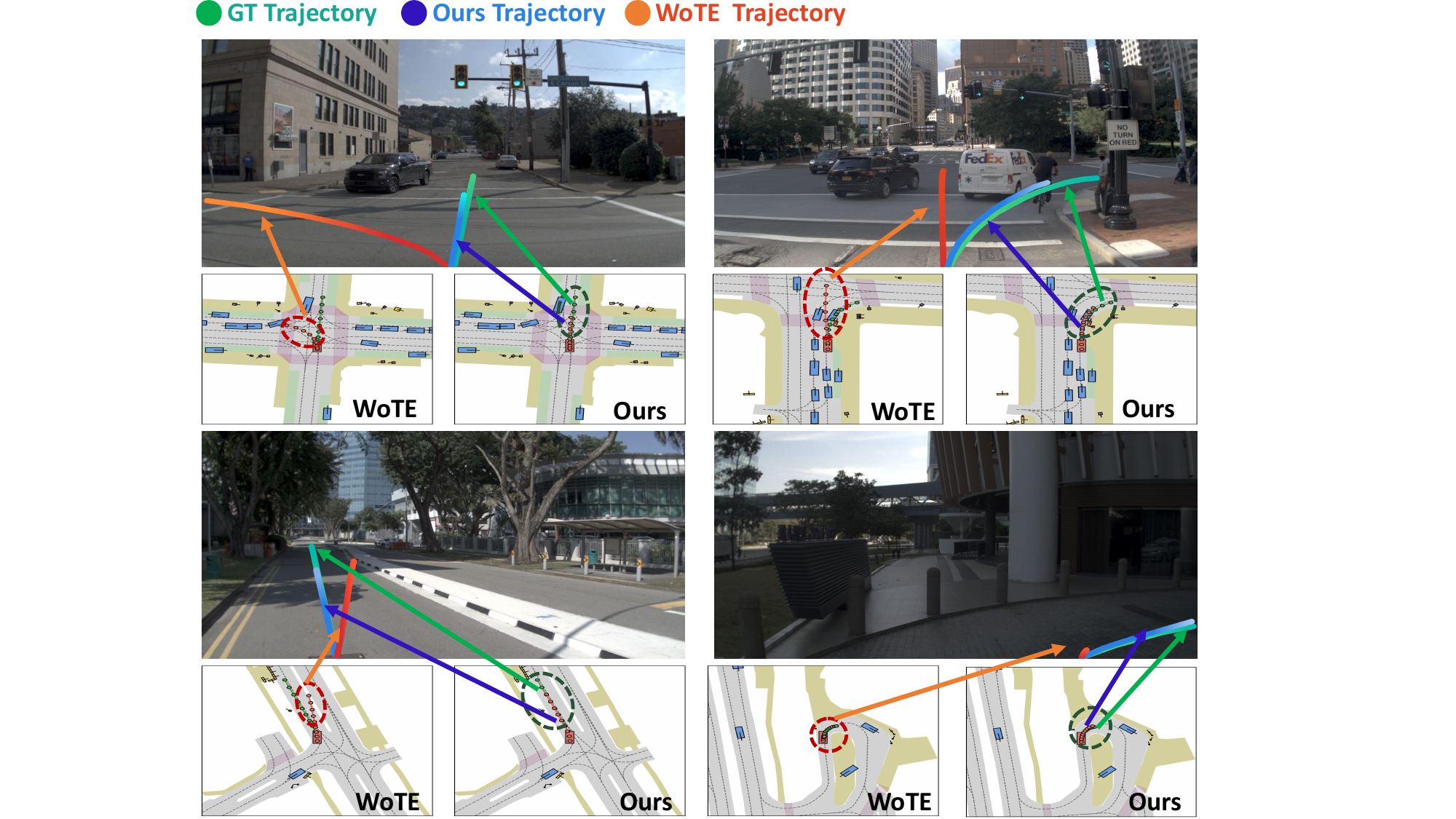}
\caption{Visualization results comparing our method with WoTE. Because our model is not constrained by trajectory anchors and learns from the full set of future features, it trains efficiently and outperforms WoTE across both simple and complex scenes, including challenging edge cases.}
\label{fig:main}
\end{figure*}
\begin{table*}[t!]
    \begin{center}
        \resizebox{0.95\textwidth}{!}{
            \begin{tabular}{l|c|c|c|cccccc}
                \toprule
                \textbf{Method} &\textbf{Input} &\textbf{\#Traj}. & \textbf{Anc/Voc}. & \textbf{NC} $\uparrow$ & \textbf{DAC} $\uparrow$ & \textbf{EP} $\uparrow$ & \textbf{TTC} $\uparrow$ & \textbf{Comf.} $\uparrow$ & \textbf{PDMS} $\uparrow$ \\ 
                \midrule
                Human & - & - & - & 100 & 100 & 87.5 & 100 & 99.9 & 94.8 \\ 
                \midrule
                VADv2~\citep{chen2024vadv2} & C & 8192 & \cmark & 97.9 & 91.7 & 77.6 & 92.9 & \bf100.0 & 83.0 \\  
                UniAD~\citep{hu2023_uniad} & C & 1 & \xmark & 97.8 & 91.9 & 78.8 & 92.9 & \bf100.0 & 83.4 \\ 
                TransFuser~\citep{chitta2022transfuser} & C \& L & 1 & \xmark & 97.7 & 92.8 & 79.2 & 92.8 & \bf100.0 & 84.0 \\
                LAW~\citep{li2024law} & C & 1 & \xmark & 96.4 & 95.4 & 81.7 & 88.7 & \underline{99.9} & 84.6 \\
                World4Drive~\citep{zheng2025world4driveendtoendautonomousdriving} & C   & 6 & \cmark & 97.4 & 94.3 & 79.9 & 92.8 & \bf100.0 & 85.1 \\
                DRAMA~\citep{yuan2024drama} & C \& L  & 1 & \xmark & 98.0 & 93.1 & 80.1 & 94.8 & \bf100.0 & 85.5 \\
                Hydra-MDP~\cite{li2024hydra} & C \& L  & 8192 & \cmark & 98.3 & 96.0 & 78.7 & 94.6 & \bf100.0 & 86.5 \\
                DiffusionDrive \cite{diffusiondrive} & C \& L  & 20 & \cmark & 98.2 & 96.2 & \underline{82.2} & \underline{94.7} & \bf100.0 & 88.1 \\
                WoTE \cite{li2025wote} & C \& L  & 256 & \cmark & \bf98.5 & \underline{96.8} & 81.6 & \bf94.8 & \underline{99.9} & \underline{88.2} \\
                \midrule
                \rowcolor{gray!20}
                Map-World (Ours) & C \& L  & 10 & \xmark & \underline{98.4} & \bf97.2 & \bf82.9 & \underline{94.7} & \bf100.0 & \bf88.8 \\
                \bottomrule
            \end{tabular}
        }
    \caption{Comparison with the SOTA approaches on NAVSIM test set. 
    Traj.: Trajectory.
    Anc/Voc.: Anchor or trajectory vocabulary.
    C: Camera.
    L: LiDAR.
    \label{tab:sota_navsim}
    }
    \end{center}
\end{table*}
\begin{table*}[t!]
    \begin{center}
        \resizebox{0.95\textwidth}{!}{
            \begin{tabular}{@{}c|ccccccccc|c@{}}
                \toprule
                \textbf{Method}& \textbf{NC}$\uparrow$ & \textbf{DAC}$\uparrow$ & \textbf{DDC}$\uparrow$ & \textbf{TL}$\uparrow$ & \textbf{EP}$\uparrow$ & \textbf{TTC}$\uparrow$ & \textbf{LK}$\uparrow$ & \textbf{HC}$\uparrow$ & \textbf{EC}$\uparrow$ & \textbf{EPDMS}$\uparrow$\\
                    \midrule
                    Human Agent &100 &100 &99.8 &100 &87.4 &100 &100 &98.1 &90.1 &90.3 \\
                    \midrule
                    Transfuser\cite{chitta2022transfuser}&96.9 &89.9 &97.8 &\underline{99.7} &87.1 &95.4 &92.7 &\bf98.3 &87.2 &76.7\\
                    VADv2\cite{chen2024vadv2}&97.3 &91.7 &77.6 &92.7 &\bf100 &\bf99.9 &\underline{98.2} &66.0 &\bf97.4 &76.6\\
                    HydraMDP\cite{li2024hydra}&97.5 &96.3 &80.1 &93.0 &\bf100 &\bf99.9 &\bf98.3 &65.5 &\bf97.4 &79.8\\
                    HydraMDP++\cite{li2025hydramdp++}&97.2 &\bf97.5 &\underline{99.4} &99.6 &83.1 &96.5 &94.4 &\underline{98.2} &70.9 &81.4\\
                    Diffusiondrive\cite{diffusiondrive}&\underline{98.0} &96.0 &\bf99.5 &\bf99.8 &87.7 &97.1 &97.2 &\bf98.3 &87.6 &\underline{84.3}\\
                    DriveSuprim\cite{yao2025drivesuprim}&97.5 & 96.5 &\underline{99.4} &99.6 &\underline{88.4} &96.6 &95.5 &\bf98.3 &77.0 &83.1\\
                    PRIX\cite{wozniak2025prix} & \underline{98.0} &95.6 &\bf99.5 &\bf99.8 &87.4 &97.2 &97.1 &\bf98.3 &87.6 &84.2\\
                    \midrule
                    \rowcolor{gray!20}
                     Map-World (Ours)& \bf98.3 & \underline{97.1} &98.9 &\bf99.8 & 87.4 & \underline{97.6} &97.2 & \bf98.3 &\underline{89.0} & \bf85.0\\
                \bottomrule
            \end{tabular}
        }
     \caption{Comparison on the NAVSIM navtest split based on closed-loop metrics. 
    \label{tab:navsimv2}
    }
    \end{center}
\end{table*}
\begin{table*} [t!] 
\centering
{\begin{tabular}[b]{l|cccc|cccc}
\toprule[1.5pt]
\multirow{2}{*}{Method} &
\multicolumn{4}{c|}{\textbf{L2} ($m$) $\downarrow$} & 
\multicolumn{4}{c}{\textbf{Col. Rate} (\%) $\downarrow$} \\
& 1$s$ & 2$s$ & 3$s$ & Avg. & 1$s$ & 2$s$ & 3$s$ & Avg. \\
\midrule 
ST-P3~\cite{hu2022stp3endtoendvisionbasedautonomous}               & 1.33 & 2.11 & 2.90 & 2.11 & 0.23 & 0.62 & 1.27 & 0.71 \\
BEV-Planner~\cite{li2024bevplanner}   & 0.28 & \bf0.42 & \bf 0.68 & \bf0.46 & 0.04 & 0.37 & 1.07 & 0.49 \\
LAW~\cite{li2024law}                  & 0.26 & 0.57 & 1.01 & 0.61 & 0.14 & 0.21 & 0.54 & 0.30 \\
PARA-Drive~\cite{weng2024paradrive}     & \underline{0.25} & \underline{0.46} & \underline{0.74} & \underline{0.48} & 0.14 & 0.23 & 0.39 & 0.25 \\
VAD-Base~\cite{jiang2023vad}             & 0.41 & 0.70 & 1.05 & 0.72 & 0.07 & 0.17 & 0.41 & 0.22 \\
GenAD~\cite{zheng2024genad}              & 0.28 & 0.49 & 0.78 & 0.52 & 0.08 & 0.14 & 0.34 & 0.19 \\
UniAD~\cite{hu2023_uniad}              & 0.44 & 0.67 & 0.96 & 0.69 & 0.04 & 0.08 & 0.23 & 0.12 \\
BridgeAD~\cite{zhang2025bridgad}        & 0.29 & 0.57 & 0.92 & 0.59 & \underline{0.01} & \underline{0.05} & 0.22 & 0.09 \\
MomAD~\cite{song2025momad}              & 0.31 & 0.57 & 0.91 & 0.60 & \underline{0.01} & \underline{0.05} & 0.22 & 0.09 \\
DiffusionDrive~\cite{diffusiondrive}  & 0.27 & 0.54 & 0.90 & 0.57 & 0.03 & \underline{0.05} & \bf0.16 & \underline{0.08} \\
\rowcolor{gray!20}
Map-World (Ours)               & \bf 0.22 & 0.47 & 0.81 & 0.50 & \bf 0.00 & \bf 0.03 & \underline{0.18} & \bf 0.07 \\
\bottomrule[1.5pt]
\end{tabular}}
\caption{
Performance comparison of planning on the nuScenes \textit{validation} split. ResNet50~\cite{he2016resnet} is used as the backbone for all methods, except for UniAD~\cite{hu2023_uniad}, which adopts ResNet101.
}
\label{tab:nuscenes}
\end{table*}

\subsection{Dataset}
\noindent\textbf{NAVSIM.}The Navsim dataset \cite{Dauner2024navsim} is a planning-oriented benchmark built on OpenScene, a 2 Hz, 120-hour condensation of nuPlan, and is resampled to emphasize challenging, non-trivial scenarios, making ego-status–only fitting insufficient. It provides 360° coverage from eight cameras plus a fused Lidar, with 2 Hz annotations including HD maps and 3D boxes. The dataset is split into Navtrain and Navtest with 1192 train/val and 136 test scenarios.

\noindent\textbf{NAVSIM metric.} The NAVSIM has two versions of metrics, one is the predictive driver model score (PDMS) \cite{Dauner2024navsim}, a weighted combination of no-accidents at-fault (NC), drivable-area compliance (DAC), time-to-collision (TTC), comfort (Comf.) and ego progress (EP).
Another is extended PDMS(EPDMS) \cite{Cao2025navsimv2}, which augments the PDMS score with four additional components: Driving Direction Compliance (DDC), Traffic Light Compliance (TLC), Lane Keeping (LK), and Extended Comfort (EC).

\noindent\textbf{nuScenes.} The nuScenes \cite{caesar2020nuscenes} dataset comprises 1000 urban driving scenes, each a 20-second synchronized multi-sensor sequence.This dataset is widely used in autonomous driving. Following the prior work \cite{jiang2023vad, sun2024sparsedriveendtoendautonomousdriving, diffusiondrive}, we report the L2 displacement Error and Collision Rate on nuScenes dataset.
\subsection{Implementation Details}
Following TransFuser and DiffusionDrive, we adopt a ResNet-34 backbone as Visual encoder. Camera inputs are \(1024\times256\) pixels. Lidar point clouds cover a \(64\text{m}\times 64\text{m}\) area around ego vehicles. We use the TransFuser backbone to produce BEV features, and the world model comprises two Transformer decoder layers. Within the world model, two BEV feature scales from the backbone are used to capture global semantics and local details. Training on Navtrain uses 2×A100-80GB GPUs with a total batch size of 256.We train for 100 epochs with a learning rate of \(6e^{-4}\). AdamW \cite{AdamW} is used as the optimizer for both datasets.
\subsection{Comparison with state of the art}
\noindent\textbf{NAVSIM.} On NAVSIM, MAP-World achieves a PDMS of 88.8 on the test set (Table~\ref{tab:sota_navsim}), surpassing all prior systems that rely on trajectory anchors or anchor vocabularies. Compared to world-model-based planners such as World4Drive and WoTE, MAP-World reaches higher PDMS without reinforcement-learning selectors or vision-language models, and improves the EPDMS score to 85.0 (Table~\ref{tab:navsimv2}). Together with the lower inference latency in Table~\ref{tab:latency}, this demonstrates that our prior-free multi-modal planning and path-weighted world model yield both better safety metrics and more efficient deployment.

\noindent\textbf{nuScenes.} On nuScenes, MAP-World also attains strong open-loop planning accuracy (Table~\ref{tab:nuscenes}). Despite the benchmark’s relative simplicity and metrics that are less sensitive to deviations from expert trajectories, our method achieves lower L2 displacement error and collision rate than most prior methods, including DiffusionDrive, while using a smaller number of modes and avoiding anchor designs.

\begin{table*} [ht!]
\centering
\vspace{-1.0em}
{\begin{tabular}{ccc|cccccc}
\toprule[1.5pt]
Mask Token & Path Integral & Traj-World &NC $\uparrow$  &DAC $\uparrow$ & TTC$\uparrow$ & Comf.$\uparrow$& EP $\uparrow$ & PDMS $\uparrow$  \\
\midrule
\cmark & \xmark &1 &\underline{98.1}&\underline{96.7}&94.3&\bf99.9&\underline{82.8}&88.4 \\
\xmark & \xmark &1& \bf98.3 &\underline{96.7} & \underline{94.7} &  \bf99.9 & 82.6 & 88.5 \\
\xmark & \cmark &4 & \bf98.3 & \underline{96.7} & \bf94.8 &  \bf99.9 & 82.3 & 88.4 \\
\cmark & \cmark &4& \bf98.3 & 96.6 & 94.4 &  \bf99.9 & 82.7 & 88.5 \\
\cmark & \cmark &8& \bf98.3 & \bf96.9 & \underline{94.7} &  \bf99.9 & 82.6 & \underline{88.6} \\
\cmark & \cmark &10& \bf98.3 & \bf96.9 &94.5  & \bf99.9 & \bf83.1 &\bf88.7  \\
\bottomrule[1.5pt]
\end{tabular}}
\caption{Ablation study on Our  Feynman Path Integral World Model. Traj-World means the number of future trajectory used in the world model. All experiments test with 2  TrajDecoder layers.}
\label{tab:masktoken}
\end{table*}
\begin{table*} [ht!]
\centering
{\begin{tabular}{ccc|cccccc}
\toprule[1.5pt]
Method & Anchor& \#Traj & NC $\uparrow$ &DAC $\uparrow$ & TTC$\uparrow$ & Comf. $\uparrow$& EP $\uparrow$ & PDMS $\uparrow$  \\
\midrule
TransFuser & \xmark &1 &97.7&92.8&92.8&\bf100&79.2&84.0 \\
DiffusionDrive (Extra)& \xmark &1 & 97.3 & 94.0 & 92.6 & \bf 100 & 79.6 & 84.7 \\
DiffusionDrive& \cmark &20&\underline{98.2} &\underline{96.2} & \bf94.7 & \bf 100 & \underline{82.2} & \underline{88.1} \\
\rowcolor{gray!20}
MAP Only (Ours)& \xmark &4& \bf98.3 & \bf96.6 & \underline{94.4} & \bf 100 & \bf82.7 & \bf88.3 \\
\bottomrule[1.5pt]
\end{tabular}}
\caption{Ablation study on the masked action planning. \#Traj means the number of future trajectory generated by model. MAP only: Only use the masked action planning.
DiffusionDrive(Extra): Denotes the trajectory extrapolated from the current state.
}
\label{tab:mode}
\vspace{-1.0em}
\end{table*}
\subsection{Ablation study.} 
\noindent\textbf{Effect of masked action planning.}
Masked action planning is pivotal,removing it causes mode collapse and unrealistic futures. To assess its effect, we drop the path-integral world model and compare with DiffusionDrive \cite{diffusiondrive} under identical settings and loss weights. As shown in Tab.~\ref{tab:mode}, DiffusionDrive (Extra) generates a single mode and yields only marginal gains over TransFuser \cite{chitta2022transfuser}, highlighting its reliance on anchors. Our module, by contrast, produces genuine multi-modal trajectories without anchors and surpasses DiffusionDrive even with anchors, confirming its effectiveness.

\noindent\textbf{Feynman Path Integral world model.} To assess the effectiveness of our Feynman path–integral world model, we conduct an ablation that isolates the entire world-model component. As shown in Tab \ref{tab:masktoken}, when conditioning on a single best trajectory to generate future states, omitting the MAE-style mask tokens yields slightly better results; however, when extending to multi-modal futures, concatenating learnable mask tokens with the current BEV features as queries produces superior performance. Moreover, under the path–integral formulation, performance improves gradually with the number of generated trajectories.

We also observe that mask tokens stabilize training as the number of modes increases, whereas removing them can lead to gradient issues. We cap the number of trajectories at 10, since beyond this point we begin to observe mode collapse in certain cases; this can be mitigated by increasing the noise perturbation in masked action planning (details in the Appendix).

\noindent\textbf{Different trajectory decoder layers.} As shown in Tab \ref{tab:decoder}, we varied the depth of the trajectory decoder and found that a 3-layer variant outperforms a 2-layer design. To balance accuracy and real-time latency, we adopt three layers for the trajectory decoder in the final model.

\begin{table}[!ht]  
    \vspace{+0.5em}  
    \begin{center}
        \resizebox{\columnwidth}{!}{
            \begin{tabular}{c|ccccc|>{\columncolor{gray!30}}c|c}
                \toprule
                Model. & \textbf{NC} $\uparrow$ & \textbf{DAC} $\uparrow$ & \textbf{EP} $\uparrow$ & \textbf{TTC} $\uparrow$ & \textbf{Comf.} $\uparrow$ & \textbf{PDMS} $\uparrow$ & \textbf{Latency} $\downarrow$\\
                \midrule
                WoTE & \bf98.5 & 96.8 & 81.6 & \bf94.8 & 99.9 & 88.2 & 18.7 ms \\
                Map-World (Ours) & 98.4 & \bf97.2 & \bf82.9 & 94.7 & \bf100.0 & \bf88.8 & \bf13.5 ms\\
                \bottomrule
            \end{tabular}
        }
    \caption{Latency test on navtest.     
    The latency is evaluated on an NVIDIA 4090 GPU.}
    \label{tab:latency}
    \end{center}
    \vspace{-1em}  
\end{table}

\begin{table}[!ht]
    \vspace{0.3em}
    \begin{center}
        \resizebox{\columnwidth}{!}{
            \begin{tabular}{c|ccccc|>{\columncolor{gray!30}}c}
                \toprule
                Traj Decoder Layer. & \textbf{NC} $\uparrow$ & \textbf{DAC} $\uparrow$ & \textbf{EP} $\uparrow$ & \textbf{TTC} $\uparrow$ & \textbf{Comf.} $\uparrow$ & \textbf{PDMS} $\uparrow$ \\
                \midrule
                2 & 98.3 & 96.9 &\bf83.1  & 94.5 & 99.9 &88.7 \\
                3 & \bf98.4 & \bf97.2 & 82.9 & \bf94.7 & \bf100.0 & \bf88.8 \\
                \bottomrule
            \end{tabular}
        }
    \caption{Ablation Study on trajectory decoder layers.}
    \label{tab:decoder}
    \end{center}
    \vspace{-1em}
\end{table}

\section{Conclusion}
\label{sec:Con}
This paper presented MAP-World, a prior-free multi-modal planning framework that couples masked action planning with a path-weighted world model for autonomous driving. By viewing planning as masked sequence completion, MAP-World generates diverse, history-consistent trajectories without handcrafted anchors or reinforcement-learning-based selectors. The path-weighted world model then supervises an expectation over candidate futures, so training benefits from the full trajectory distribution instead of a single chosen mode. Experiments on NAVSIM and nuScenes show that MAP-World improves both safety-critical driving metrics and planning accuracy, while keeping inference latency compatible with real-time deployment. A current limitation is the fixed number of trajectory modes and the sensitivity of training to probability weighting; future work will explore scene-adaptive mode allocation and more robust weighting schemes.These will all be the focus of our future work, and further research will be conducted.

\noindent\textbf{Acknowledgements.} This work was supported by the Science and Technology Development Fund of Macau [0122/2024/RIB2, 0215/2024/AGJ, 001/2024/SKL], the Research Services and Knowledge Transfer Office, University of Macau [SRG2023-00037-IOTSC, MYRG-GRG2024-00284-IOTSC], the Shenzhen-Hong Kong-Macau Science and Technology Program Category C [SGDX20230821095159012], the Science and Technology Planning Project of Guangdong [2025A0505010016], National Natural Science Foundation of China [52572354], the State Key Lab of Intelligent Transportation System [2024-B001], and the Jiangsu Provincial Science and Technology Program [BZ2024055].
{
    \small
    \bibliographystyle{ieeenat_fullname}
    \bibliography{main}
}

\clearpage
\setcounter{page}{1}
\maketitlesupplementary
\appendix

\section{Further Ablation study}
We conducted further ablation study on the number of BEV states predicted by our world model to investigate how the granularity of prediction time steps affects final performance. This experiment was conducted using two GeForce RTX 3090 GPU with a learning rate of $2 \times 10^{-4}$. As shown in Tab \ref{tab:ts}, employing finer, denser time steps to predict future states increased model complexity yet resulted in a marginal decline in overall performance. Specifically, the configuration predicting only the fourth-second BEV achieved higher scores than simultaneously predicting both the 2s and 4s BEVs.
\begin{table}[!ht]
    \vspace{0.3em}
    \begin{center}
        \resizebox{\columnwidth}{!}{
            \begin{tabular}{ccc|ccccc|>{\columncolor{gray!30}}c}
                \toprule
                Prediction Steps &  \#Traj& Traj Decoder Layer.&\textbf{NC} $\uparrow$ & \textbf{DAC} $\uparrow$ & \textbf{EP} $\uparrow$ & \textbf{TTC} $\uparrow$ & \textbf{Comf.} $\uparrow$ & \textbf{PDMS} $\uparrow$ \\
                \midrule
                0s→(2s,4s) &8 &2 &\bf98.3 & 96.2 &82.3  & \bf94.7 & 100.0 &88.1 \\
                0s→4s & 8 & 2&\bf98.3 & \bf96.7 & \bf82.6 & 94.6 & \bf100.0 & \bf88.4 \\
                \bottomrule
            \end{tabular}
        }
    \caption{Ablation Study on future prediction steps. A→(B,C): predicted future states at time B and C are based on the state at time A. \#Traj: The number of future trajectory generated by model.}
    \label{tab:ts}
    \end{center}
    \vspace{-1em}
\end{table} 

\section{Various noise perturbation factors}
\begin{figure*}[t!]
\centering
\includegraphics[width=1\textwidth]{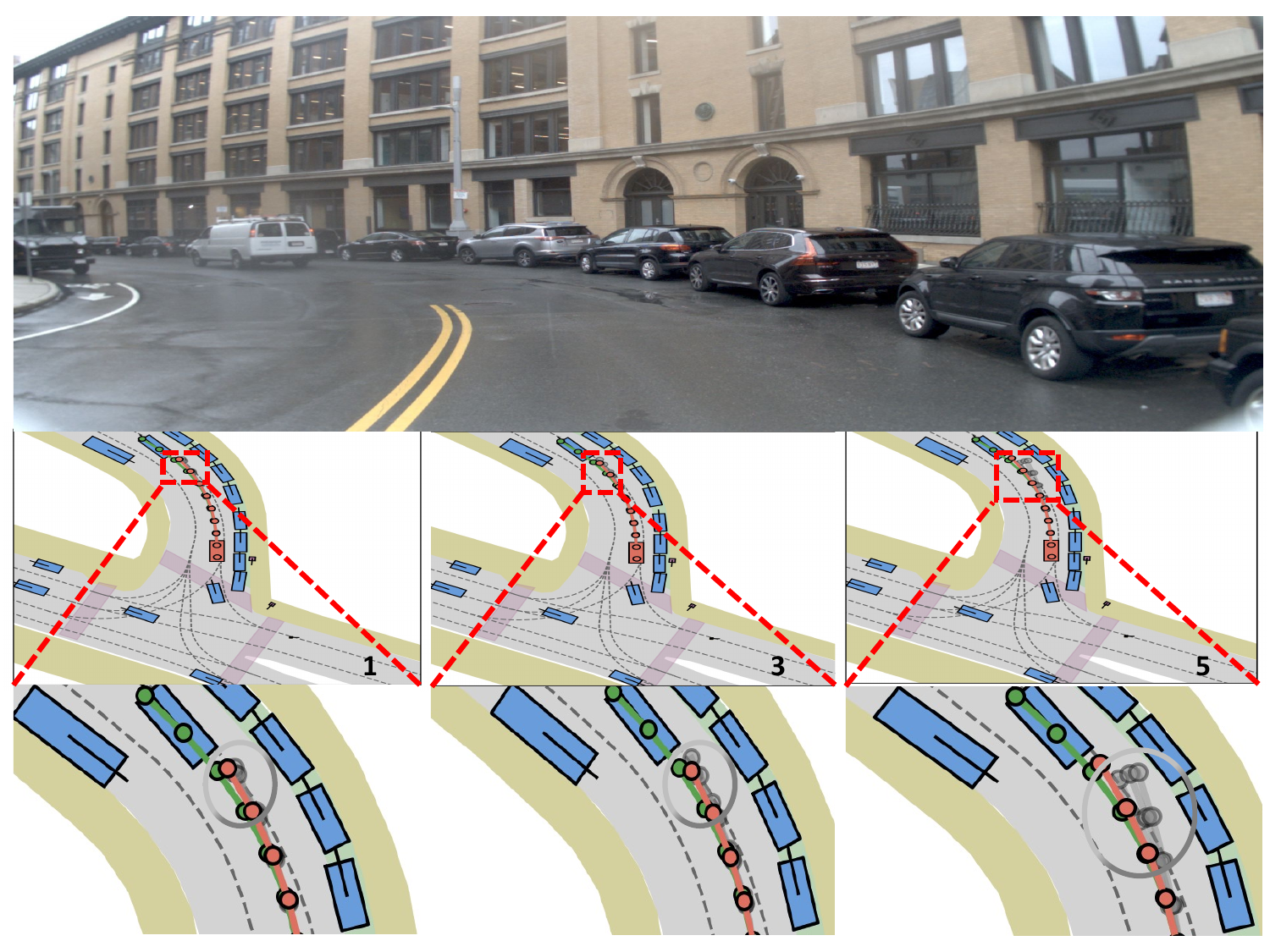}
\caption{Visualization of trajectories with noise perturbations of different factors.
}
\label{fig:noise}
\end{figure*}

 When the number of generated trajectories reaches ten, the resulting modes become visually entangled and difficult to distinguish. Amplifying the noise perturbation used in masked action planning by different scaling factors can make the multi-modal trajectories more separable in visualization, as shown in Figure \ref{fig:noise}, but it also degrades planning quality, as shown in Tab \ref{tab:noise}.  This indicates that, although the proposed paradigm is effective, additional research is needed to improve robustness under stronger noise perturbations.Therefore,we do not recommend training with excessively large noise perturbations.

 \begin{table}[!ht]
    \vspace{0.3em}
    \begin{center}
        \resizebox{\columnwidth}{!}{
            \begin{tabular}{c|ccccc|>{\columncolor{gray!30}}c}
                \toprule
                Noise Factor &\textbf{NC} $\uparrow$ & \textbf{DAC} $\uparrow$ & \textbf{EP} $\uparrow$ & \textbf{TTC} $\uparrow$ & \textbf{Comf.} $\uparrow$ & \textbf{PDMS} $\uparrow$ \\
                \midrule
                1 &\bf98.4 & \bf97.2 &\bf82.9 & \bf94.7 & \bf100.0 &\bf88.8 \\
                3 & 98.3 & \bf97.2 & 82.8 & 94.5 & \bf100.0 & 88.7 \\
                5 & 98.3 & 97.0 & 82.6 & 94.5 & \bf100.0 & 88.5 \\
                \bottomrule
            \end{tabular}
        }
    \caption{Ablation Study on noise perturbation factors. }
    \label{tab:noise}
    \end{center}
    \vspace{-1em}
\end{table}

\section{Further Model Training Detail}
In this section, we compare the data requirements of our approach with those of prior methods. As shown in the Tab \ref{tab:sota_qualitative}, within purely end-to-end planning frameworks, anchor-based methods significantly outperform anchor-free approaches on the NAVSIM test dataset. For world-model–based methods, using additional data beyond standard end-to-end planning inputs is common. For example, WoTE not only utilizes the anchor trajectory but also relies on extra annotations by inserting an ego box into the future BEV semantic map that does not appear in the original dataset to force alignment between the predicted future trajectory and the annotated ego box. This provides a stronger supervisory signal for the BEV world model and enhances its understanding of physical context. In contrast, our method does not use any such additional annotations, with only simple historical trajectory inputs, it surpasses WoTE, further demonstrating the effectiveness of our approach.
\begin{table*}[t!]
    \begin{center}
        \resizebox{0.95\textwidth}{!}{
        \begin{tabular}{lccccccc}
        \toprule
        \textbf{Method} & \textbf{Type} &
        \makecell{\textbf{Multiple}\\\textbf{Sensors}} &
        \makecell{\textbf{Ego}\\\textbf{Status}} &
        \makecell{\textbf{Hist}\\\textbf{Traj.}} &
        \makecell{\textbf{Ego}\\\textbf{Box}} &
        \makecell{\textbf{Voc \&}\\\textbf{Anchors}} &
        \textbf{PDMS} $\uparrow$ \\
        \midrule
        UniAD & E2E Plan  & \cmark & \cmark & \xmark & \xmark & \xmark & 83.4 \\
        Transfuser & E2E Plan & \cmark & \cmark & \xmark & \xmark & \xmark & 84.0 \\
        DiffusionDrive & E2E Plan &
        \cmark & \cmark & \xmark & \xmark & \cmark & 88.1 \\
        \midrule
        DrivingGPT\cite{chen2024drivinggpt} & WM + E2E Plan &
        \xmark & \xmark & \cmark & \xmark & \xmark & 82.4 \\
        LAW & WM + E2E Plan & \cmark & \cmark & \xmark & \xmark & \xmark & 84.6 \\
        WoTE & WM + E2E Plan & \cmark & \cmark & \xmark & \cmark & \cmark & 88.2 \\
        \midrule
        \textbf{Map-World (Ours)} & WM + E2E Plan &
        \cmark & \cmark & \cmark & \xmark & \xmark & \textbf{88.8} \\
        \bottomrule
        \end{tabular}
        }
    \caption{Comparison on the information utilized in model training. E2E Plan: End-to-End planning. WM: World Model. Hist Traj: History Trajectory. Voc\&Anchors: trajectory vocabulary or trajectory anchors.} 
    \label{tab:sota_qualitative}
    \end{center}
\end{table*}

\section{Further Qualitative Comparison}
In this section, we further present qualitative comparisons between our method and WoTE on challenging scenarios from the NAVSIM test dataset.

\begin{figure*}[t!]
\centering
\includegraphics[width=1\textwidth]{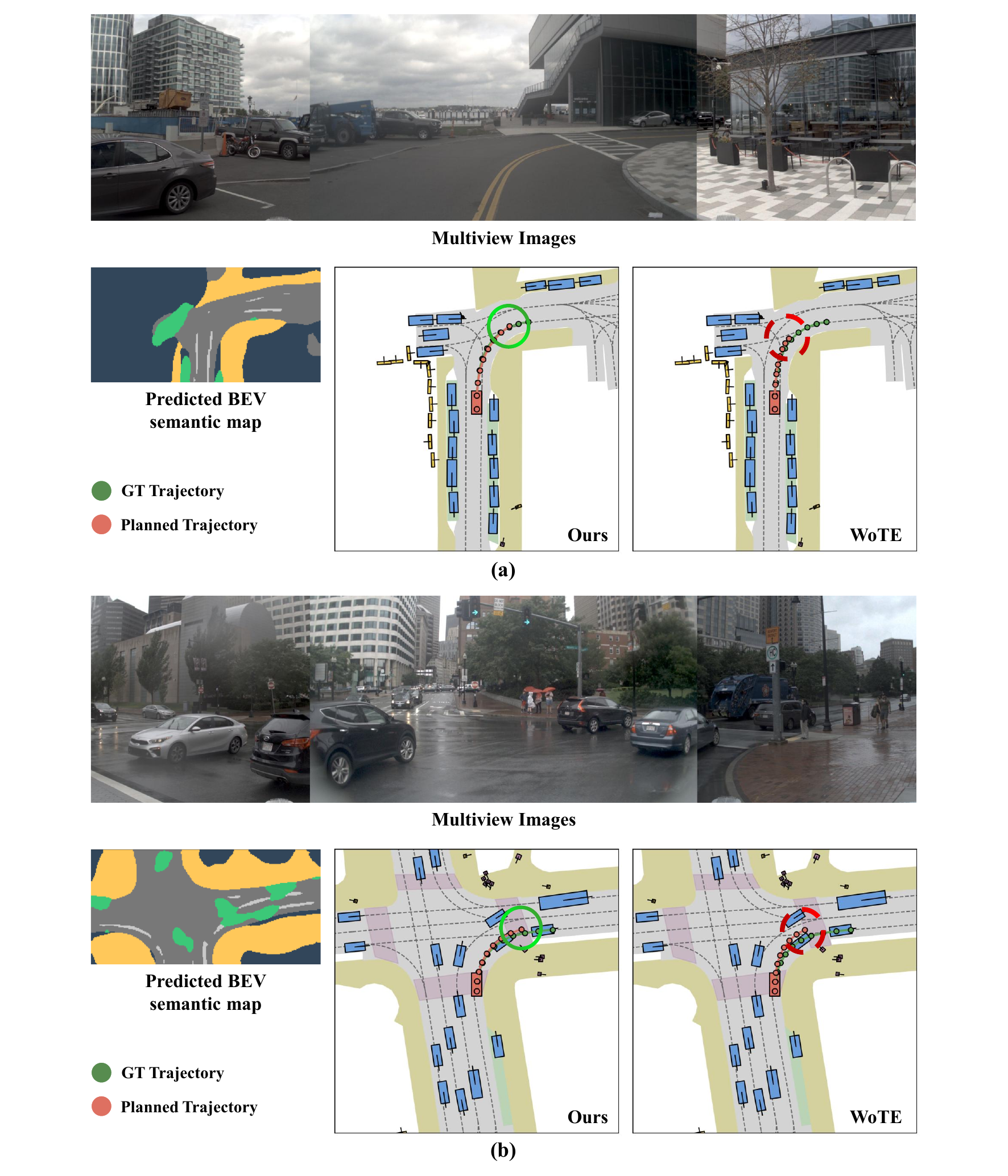}
\caption{Qualitative comparison of WoTE and Map-World on turning right scenarios of NAVSIM navtest split.}
\label{fig:right}
\end{figure*}

\noindent\textbf{Turning right.} The Figure \ref{fig:right} shows that, compared with WoTE, MAP-World closely follows the ground-truth trajectory in complex right-turn scenarios while maintaining a high throughput, completing the maneuver more quickly. In contrast, WoTE tends to exhibit pronounced hesitation or produces trajectories with larger turning radii that deviate from typical human expert driving behavior.

\begin{figure*}[t!]
\centering
\includegraphics[width=1\textwidth]{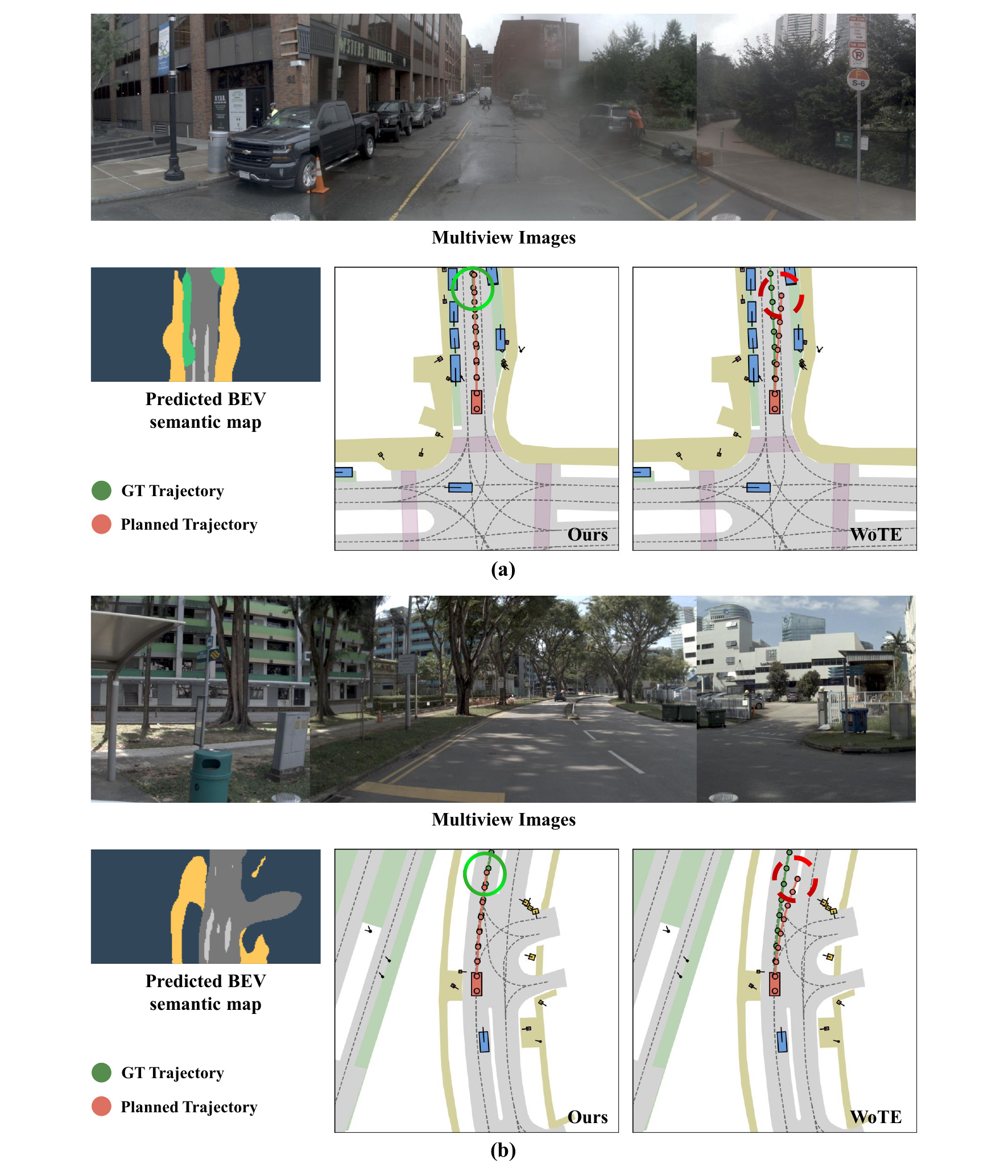}
\caption{Qualitative comparison of WoTE and Map-World on going straight scenarios of NAVSIM navtest split.}
\label{fig:straight}
\end{figure*}

\noindent\textbf{Going straight.} The Figure \ref{fig:straight} shows that, in contrast, MAP-World maintains stable forward motion and robust lane-keeping even on narrow roads and in tight-radius curves, whereas WoTE is more prone to lane departures and shows reduced longitudinal stability in such scenarios.

\begin{figure*}[t!]
\centering
\includegraphics[width=1\textwidth]{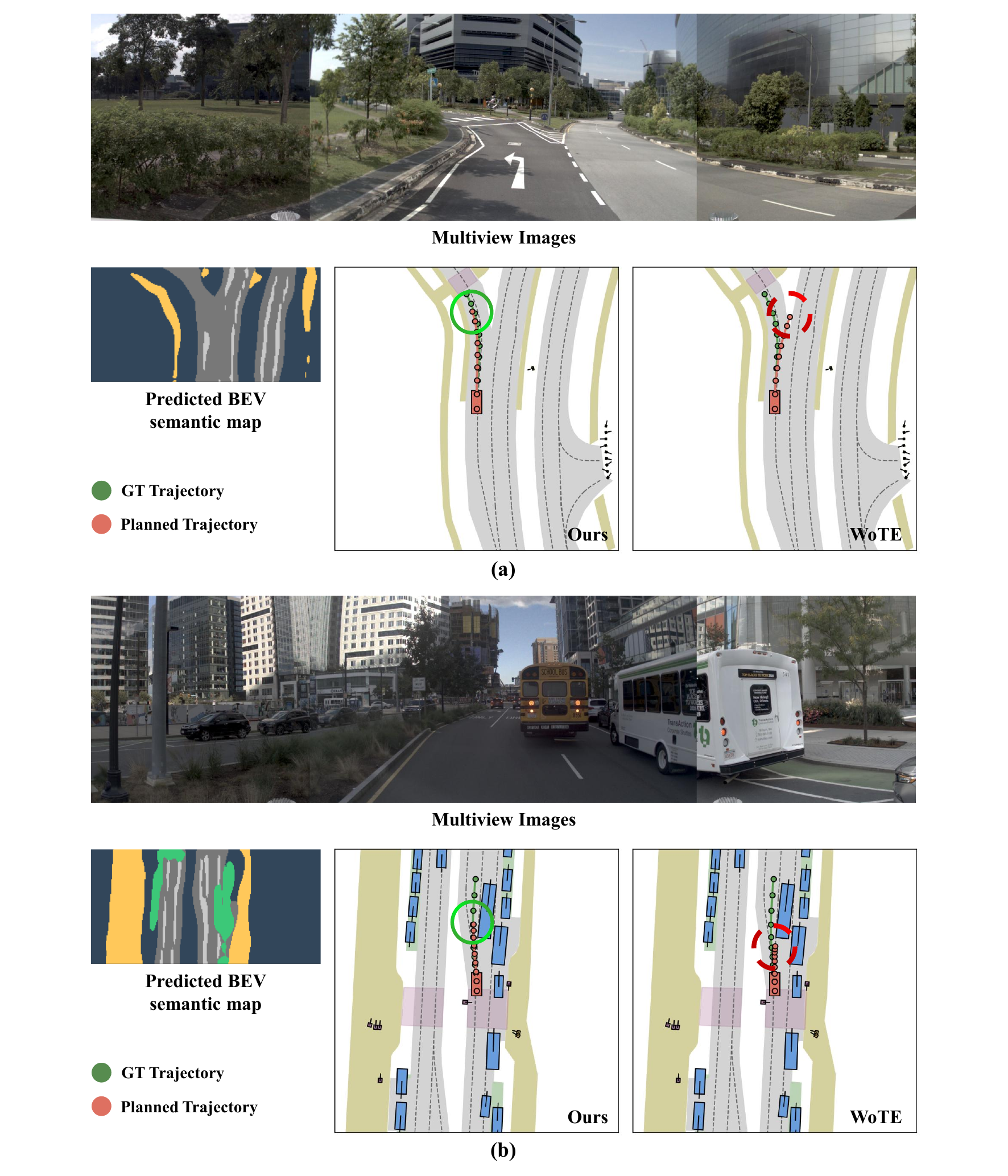}
\caption{Qualitative comparison of WoTE and Map-World on turning left and overtaking scenarios of NAVSIM navtest split.}
\label{fig:left}
\end{figure*}

\noindent\textbf{Turning left and overtaking.} The Figure \ref{fig:left} shows that,
in some scenarios, WoTE tends to misinterpret turning intent, for example classifying a left turn as a right turn, which leads to planned trajectories that deviate markedly from the desired route. In overtaking scenarios, when a large vehicle (e.g., a bus) occludes the field of view, WoTE often adopts an overly conservative strategy and fails to initiate an overtake. This behavior is likely due to the scarcity of such cases in the training data, so that the corresponding trajectory patterns are not well represented in the anchor set, making it difficult for the model to learn appropriate decisions for these situations.

\end{document}